\pdfoutput=1
\documentclass[11pt]{article}

\usepackage{ACL2023}

\usepackage{times}
\usepackage{latexsym}

\usepackage[T1]{fontenc}
\usepackage[utf8]{inputenc}

\usepackage{microtype}

\usepackage{inconsolata}

\usepackage{makecell}
\usepackage{graphicx}
\usepackage{amssymb}
\usepackage{multirow}
\definecolor{Gray}{gray}{0.9}
\usepackage{color, colortbl}
\title{Boosting Radiology Report Generation by Infusing Comparison Prior}

\author{Sanghwan Kim $^{\clubsuit \heartsuit}$
\quad Farhad Nooralahzadeh$^{\heartsuit}$ \quad Morteza Rohanian $^{\heartsuit}$ \\ {\bf Koji Fujimoto}$^{\diamondsuit}$ \quad {\bf Mizuho Nishio}$^{\diamondsuit}$\quad {\bf Ryo Sakamoto}$^{\diamondsuit}$\\ {\bf Fabio Rinaldi}$^{\spadesuit}$\quad {\bf Michael Krauthammer}$^{\heartsuit}$\\
$^{\clubsuit}$ ETH Zürich\\
$^{\heartsuit}$University of Zürich and
University Hospital of Zürich\\
     $^{\diamondsuit}$Kyoto University Graduate School of Medicine\\
     $^{\spadesuit}$Dalle Molle Institute for Artificial Intelligence Research\\
    \texttt{sanghwan.kim@inf.ethz.ch}
}

\begin{document}
\maketitle
\begin{abstract}
Recent transformer-based models have made significant strides in generating radiology reports from chest X-ray images. However, a prominent challenge remains: these models often lack prior knowledge, resulting in the generation of synthetic reports that mistakenly reference non-existent prior exams. This discrepancy can be attributed to a knowledge gap between radiologists and the generation models. While radiologists possess patient-specific prior information, the models solely receive X-ray images at a specific time point. To tackle this issue, we propose a novel approach that leverages a rule-based labeler to extract comparison prior information from radiology reports. This extracted comparison prior is then seamlessly integrated into state-of-the-art transformer-based models, enabling them to produce more realistic and comprehensive reports. Our method is evaluated on English report datasets, such as IU X-ray and MIMIC-CXR. The results demonstrate that our approach surpasses baseline models in terms of natural language generation metrics. Notably, our model generates reports that are free from false references to non-existent prior exams, setting it apart from previous models. By addressing this limitation, our approach represents a significant step towards bridging the gap between radiologists and generation models in the domain of medical report generation.
\end{abstract}

\section{Introduction}

\begin{table*}
\centering
\begin{tabular}{p{0.3\linewidth}  p{0.3\linewidth}  p{0.3\linewidth}}
\hline
\multicolumn{1}{c}{Ground Truth} & \multicolumn{1}{c}{R2Gen \cite{chen2020generating}} & \multicolumn{1}{c}{$M^2$Tr \cite{cornia2020meshed}}\\ \hline
{\small Heart size is normal. Aorta is tortuous and ectatic. Cardiomediastinal contours are normal. Lungs are clear without evidence of fibrosis. Pleural effusions or pneumothorax. Endplate sclerotic changes are present in the thoracic spine.}  & {\small There are diffuse increased interstitial suggestive of pulmonary fibrosis in bilateral lung xxxx. The fibrosis appears to slightly increased xxxx \textbf{compared to previous} in xxxx. The trachea is midline. negative for pleural effusion. the heart size is normal.}   & {\small Both lungs are clear and expanded. Heart and mediastinum normal.}       \\ \hline

{\small Stable cardiomediastinal silhouette. No focal pulmonary pleural effusion or pneumothorax. No acute bony abnormality.}  & {\small The heart is normal size. The mediastinum is unremarkable. There is no pleural or focal airspace disease. Mild chronic degenerative changes are present in the spine.}   & {\small Low lung volumes. Elevation of the right hemidiaphragm. Patchy opacities right base \textbf{again noted}. Left lung clear. Heart size top normal. Aortic calcification. Granulomas. No evidence of pneumothorax. Blunting of the bilateral costophrenic xxxx. Degenerative changes of the thoracic spine .}       \\ \hline
\end{tabular}
\caption{Ground truth report from IU X-ray (first column) and examples of reports generated by R2Gen and $M^2$Tr (second and third column). Prior expressions are emphasized in bold.}
\label{tab:report_examples}
\end{table*}

The analysis of radiology images and the subsequent writing of medical reports are crucial tasks performed during the diagnostic process \cite{suetens2017fundamentals, krupinski2010current}. However, producing a radiology report is a labor-intensive and time-consuming task for radiologists, requiring years of training to accurately identify and describe specific abnormalities in medical images \cite{brady2017error, arenson2006training}. Inspired by the success of image captioning models in deep learning, numerous studies have emerged proposing various models for automated radiology report generation, specifically focusing on chest X-ray images \cite{yuan2019automatic, li2019knowledge, xue2018multimodal, jing2017automatic, liu2019clinically}. The automated generation of reports holds the potential to alleviate the high workload of radiologists and expedite the diagnostic process by providing preliminary reports that include useful keywords or observations \cite{johnson2019mimic, chen2020generating}.

Despite the relative success of recent approaches in generating radiology reports from chest X-ray images \cite{endo2021retrieval, johnson2019mimic, chen2020generating, miura2020improving, ramirez2022medical, nooralahzadeh2021progressive}, a crucial challenge remains unaddressed in these studies: the need to provide models with appropriate prior knowledge, akin to what is available to radiologists. Specifically, radiologists are equipped with information about the existence of previous reports and X-ray images, enabling them to compare current exams with past ones, and assess the patient's progress, deterioration, or improvement \cite{suetens2017fundamentals, european2011good}. These medical reports often incorporate specific words or phrases for comparison, such as "compared to the previous exam," "in the interval," "referring to the prior X-ray," and so on. In this paper, we refer to these words or phrases as \emph{prior expressions}, which are also present in general medical datasets such as MIMIC-CXR \cite{johnson2019mimic} and IU X-ray \cite{demner2016preparing}, widely utilized for training and evaluating report generation tasks. The inclusion of prior expressions in medical reports is vital for accurate reporting. However, models trained on medical report datasets often generate reports with inappropriate or misused prior expressions, leading to relatively lower performance metrics. The challenge lies in effectively incorporating and utilizing prior expressions within the model's generation process, a crucial aspect yet to be fully resolved.

In Table~\ref{tab:report_examples}, we present a comparison between ground truth reports and reports generated by two recent models, R2Gen \cite{chen2020generating} and $M^2$Tr \cite{cornia2020meshed}, focusing on the presence of prior expressions. It is evident that the synthetic reports contain inappropriate priors. For instance, the synthetic report generated by R2Gen in the first row includes a comparison with the previous exam, despite the absence of any prior information in the ground truth report. Similarly, the synthetic report produced by $M^2$Tr in the second row includes the phrase "again noted," indicating the existence of a previous image, while the ground truth report lacks any prior expression.

\begin{figure}
\centering
\includegraphics[scale=.5]{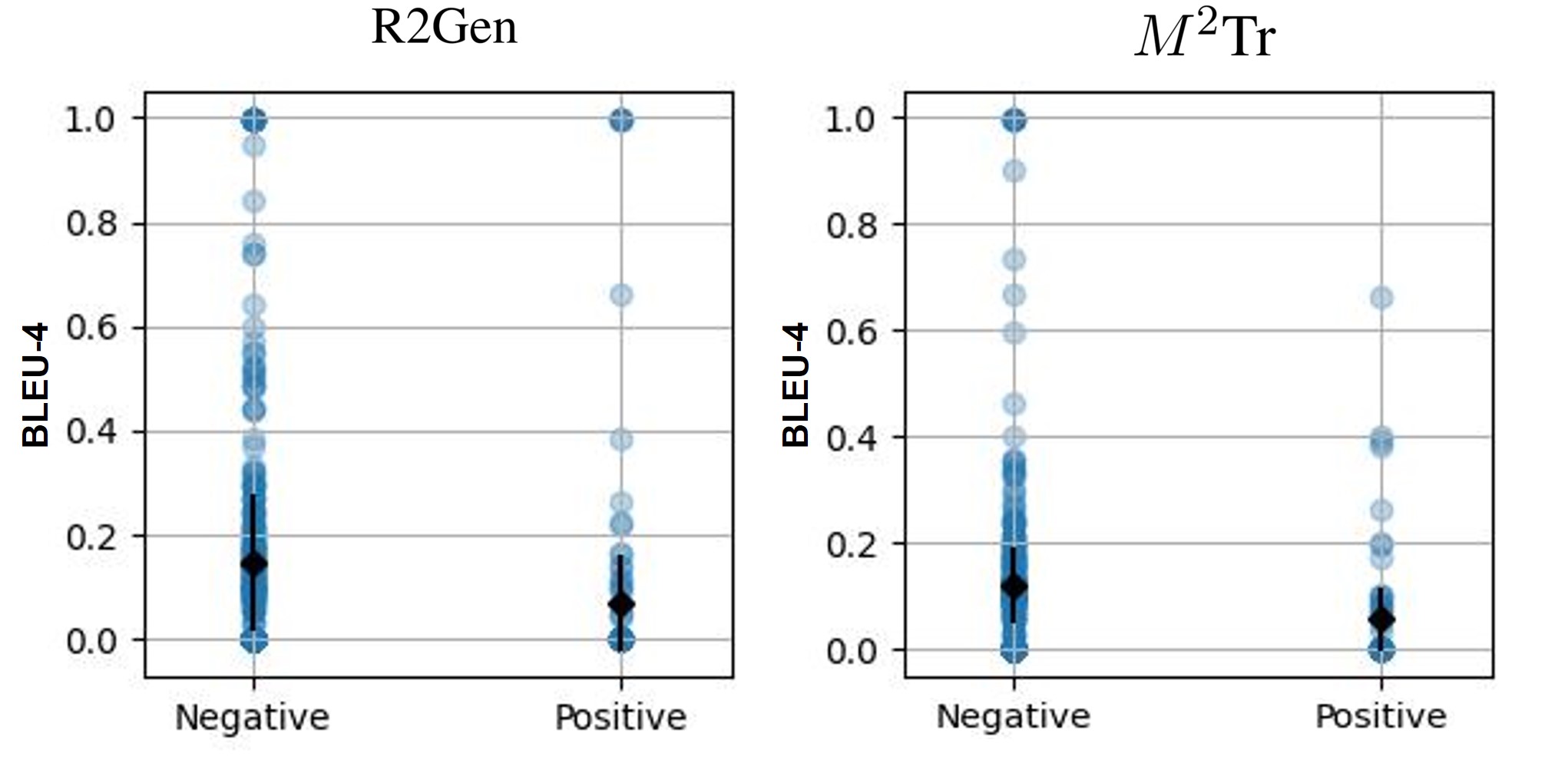}
\caption{The distribution of BLEU-4 scores based on the classification by our rule-based labeler into negative and positive categories. The labeler evaluated synthetic reports generated by R2Gen and $M^2$Tr, both trained on the IU X-ray dataset. The mean and standard deviation are represented by black dots and lines, respectively. The mean scores for negative and positive labels are as follows: for R2Gen, it is (0.1455, 0.0698), and for $M^2$Tr, it is (0.1218, 0.0583).}
\label{fig:labeler_score_distribution}
\end{figure}

These falsely referenced reports, which include prior expressions, tend to yield lower evaluation metrics. In Figure~\ref{fig:labeler_score_distribution}, we utilize a rule-based labeler (explained in Section~\ref{subsec:labeler}) to classify synthetic reports into two categories: negative and positive. The negative class represents reports without prior expressions, while the positive class comprises reports with prior expressions. We then plot the distribution of BLEU-4 \cite{papineni2002bleu} scores for each class, along with their respective mean and standard deviation. The results clearly demonstrate that the positive reports achieve lower scores compared to the negative reports. This observation suggests that the transformer-based models do not effectively leverage the generated prior expressions, underscoring the significance of properly incorporating prior information during the training process.

Table~\ref{tab:report_examples} and Figure~\ref{fig:labeler_score_distribution} highlight a critical issue with state-of-the-art (SOTA) models in learning and generating prior expressions, which can potentially confuse radiologists when utilizing these models for report writing. This issue stems from the fundamental disparity between how radiologists compose reports and how models generate them. Radiologists have access to not only prior patient information such as previous exams and medical history, but also the current X-ray images. In contrast, report generation models are only provided with X-ray images at a specific moment. With limited prior information, it becomes challenging for the models to generate comprehensive and insightful reports comparable to those created by experts.

In our paper, we address this knowledge gap between generation models and radiologists by infusing prior information into existing models. Our aim is to reduce the disparity and enable the improved models to produce more informative and practical reports. Since existing datasets such as IU X-ray and MIMIC-CXR do not contain prior information (previous X-ray images), we adopt a data-driven approach by consulting experienced radiologists. Inspired by the CheXpert labeler \cite{irvin2019chexpert}, we develop a rule-based labeler that extracts prior information by identifying specific patterns and keywords in radiology reports. Notably, the rule-based labeler focuses on comparison phrases that indicate whether a medical report corresponds to the first or subsequent exams for each patient.

The main contributions of our paper are as follows:
\begin{enumerate}
    \item We collaborate with radiologists to develop a rule-based labeler that identifies specific keywords and patterns in medical reports related to comparisons.
    \item To incorporate prior information into the models, we propose an enhanced transformer-based architecture. This approach is straightforward to implement and can be seamlessly integrated as a plug-in method into modern report generation models.
    \item Through empirical evaluation on the IU X-ray and MIMIC-CXR datasets, we demonstrate that our model outperforms baseline models in terms of performance metrics.
    \item Furthermore, we conduct a comprehensive analysis to confirm that our model no longer generates false references, addressing a significant limitation observed in previous approaches.
\end{enumerate}

\section{Related Work}
Initial research \cite{bai2018survey, liu2019survey} in radiology report generation employed a basic encoder-decoder architecture, where an encoder extracted key features from medical images and converted them into a latent vector, and a decoder generated the target text from the latent vector. Typically, CNN \cite{lecun2015deep} was used as the encoder, and LSTM \cite{hochreiter1997long} was chosen as the decoder. Subsequently, visual attention mechanisms were introduced to highlight specific image features and generate more interpretable reports \cite{zhang2017mdnet, jing2017automatic, wang2018tienet, yin2019automatic, yuan2019automatic}. Recent studies \cite{lovelace2020learning, chen2020generating, nooralahzadeh2021progressive, miura2020improving} have explored more advanced architectures using transformers to produce more comprehensive and consistent medical reports.

Alternatively, generating medical reports can be approached as a retrieval task, as similar sentences and a specific writing format are often repeated in most reports. Reusing diagnostic text from visually similar X-ray images may result in more consistent and accurate reports compared to generating an entire report from scratch. \citet{li2019knowledge} demonstrated the superiority of retrieval-based models, outperforming many encoder-decoder-based models. More recently, \citet{endo2021retrieval} introduced a retrieval-based model called CXR-RePaiR, which incorporated contrastive language image pre-training (CLIP) \cite{radford2021learning} to calculate the similarity between text and image embeddings. CXR-RePaiR generates predictions by selecting the most aligned report from a large report corpus given a specific X-ray image. They achieved SOTA performance on their newly developed metrics.

Previous approaches have primarily focused on enhancing model performance through advancements in model architecture, while paying relatively less attention to the distinctive characteristics of radiology reports, particularly those involving comparisons. However, there are a few notable exceptions, such as the work by \citet{ramesh2022improving}, which specifically explores the impact of prior expressions in reports. The authors of that study introduced a novel dataset named MIMIC-PRO, in which they identified and modified reports that contained hallucinated references to non-existent prior exams. The hallucinated references can be seen as a concept analogous to the prior expressions discussed in our paper. \citet{ramesh2022improving} suggested a BioBERT-based model that paraphrased or removed sentences referring to previous reports or images, arguing that these expressions confuse the model and result in falsely referenced sentences. They collaborated with experts to create a "clean" MIMIC-CXR test dataset and compared models trained on MIMIC-CXR and MIMIC-PRO. 

In contrast to \citet{ramesh2022improving}, we take a different approach to address the issue of comparison priors: we include prior information in the model and enable it to generate more comprehensive reports in an end-to-end fashion, rather than entirely removing comparison priors. Writing comparisons using prior expressions in radiology reports is unavoidable in the real medical field, and constructing a clean and accurate dataset from real reports is also a laborious task. Thus, our work focused on directly applying the comparison prior to existing models such as R2Gen and $M^2$Tr.

\section{Method}
\begin{table}
\centering
\begin{tabular}{p{0.7\linewidth}  p{0.2\linewidth}}
\hline
\multicolumn{1}{c}{Report} & \multicolumn{1}{c}{Label} \\ \hline
{\small 1. Cardiomegaly is noted and is stable \textbf{compared to prior examination} from XXXX.}   &  \multicolumn{1}{c}{1}     \\
 {\small 2. Ill-defined opacity is \textbf{again noted} in the region of the lingula.}   &  \multicolumn{1}{c}{1}     \\
 {\small 3. There are low lung volumes. The lungs are otherwise clear.}     &  \multicolumn{1}{c}{0}     \\
{\small  4. The left lower lobe have cleared \textbf{in the interval.}}      &  \multicolumn{1}{c}{1}    \\ \hline
\end{tabular}
\caption{Output of the labeler given sampled reports from the IU X-ray dataset. The bolded phrases represent the prior expressions identified by our labeler.}
\label{tab:examples_of_labeler}
\end{table}

\begin{table}
\begin{tabular}{cccc}
\hline
Dataset   & Negative & Positive & Total \\ \hline
IU X-ray  &  3,426    & 529         &  3,955     \\
MIMIC-CXR &  106,628 & 99,935      &  206,563   \\ \hline
\end{tabular}
\caption{Number of studies classified as negative (0) or positive (1) by our rule-based labeler.}
\label{tab:output_dist_of_labeler}
\end{table}

\begin{figure*}
\centering
\includegraphics[scale=.55]{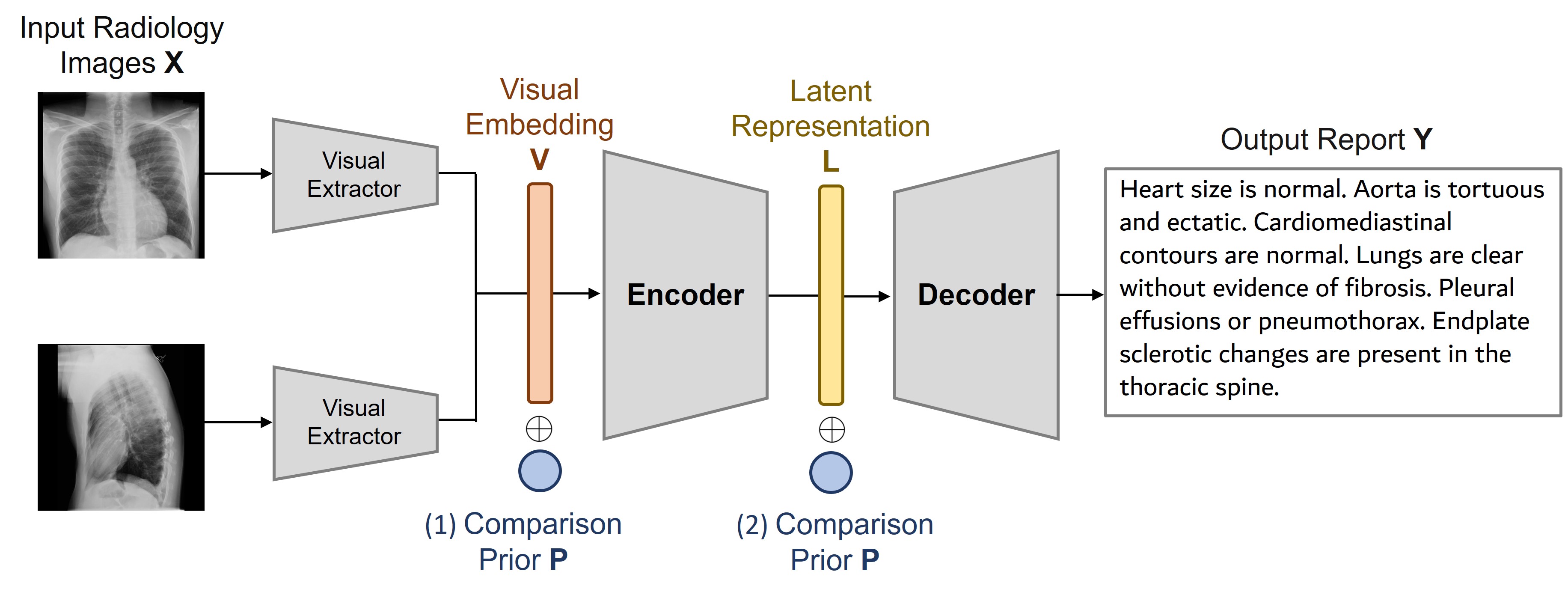}
\caption{A conceptual diagram of our approach. The report generation models (R2Gen and $M^2$ Tr) consist of a Visual Extractor, Encoder, and Decoder. Our key idea is to infuse comparison priors generated by our rule-based labeler into (1) Visual Embedding $V$ and (2) Latent Representation $L$.}
\label{fig:concept}
\end{figure*}

As observed in Table~\ref{tab:report_examples}, even the best models to date struggle with unexpected prior expressions. To address this problem, we propose a two-step approach: (1) constructing a rule-based labeler to differentiate reports with and without prior expressions (Section~\ref{subsec:labeler}), and (2) extending the transformer-based architectures (R2Gen and $M^2$Tr) to incorporate the comparison prior as input (Section~\ref{subsec:ext_model}). In the first step, we introduce a rule-based labeler that detects specific comparison prior expressions and categorizes each report as either a first exam (negative) or a following exam (positive), based on its detection. We draw inspiration from the negation and classification principles of the CheXpert labeler \cite{irvin2019chexpert} to design our novel labeler. In the subsequent stage, we integrate our prior label as input into the state-of-the-art architectures to generate more practical and comprehensive reports, and we compare the results with the baseline models. By providing our novel model with the comparison prior, which is typically communicated to radiologists in real diagnostic scenarios, we enable the generation of more comprehensive and consistent medical reports.

\subsection{Rule-based Labeler}\label{subsec:labeler}
Our rule-based labeler follows the fundamental structure of the CheXpert labeler \cite{irvin2019chexpert}, which detects the presence of 14 observations in radiology reports based on fixed rules devised by experts. Consequently, our labeler consists of three distinct stages: mention extraction, mention classification, and mention aggregation. The labeler takes the Finding section of radiology reports as input and generates a binary output (0 or 1). A negative label (0) indicates a report without prior expressions, while a positive label (1) signifies the presence of prior expressions.

\paragraph{Mention extraction}
A mention is defined as a specific keyword likely to be included in prior expressions, such as "previous", "prior", "preceding", "previously", "again", "comparison", "interval", "increase", "decrease", "enlarge", and so on. In this stage, the labeler extracts mentions from each report and marks them within each sentence. However, it is important to note that even if certain sentences contain the designated keywords, the existence of a prior expression cannot be confirmed at this step since those keywords might be used in other contexts. For instance, the word "comparison" can be used in a sentence like "with no comparison studies," indicating the absence of prior expressions.
 
\paragraph{Mention classification}
After extracting mentions in the first stage, our labeler determines whether each mention corresponds to predefined prior expressions. As similar expressions are frequently employed in reports to denote a comparison with previous exams, we can formalize the patterns of these prior expressions into several key phrases familiar to experienced radiologists. For example, the phrase "compared / similar to \{mention\}" confirms the presence of prior reports, where "\{mention\}" represents keywords such as "previous", "preceding", and "prior". Similarly, "\{mention\} seen/identified/visualized/ ... /noted" constitutes a prior expression when "\{mention\}" pertains to keywords like "again" and "previously".

\paragraph{Mention aggregation}
In the final stage, the labeler combines the classified mentions and generates either a negative label (0) or a positive label (1), with negative indicating a report without prior expressions and positive denoting the presence of prior expressions. Examples of the labeler's outputs can be seen in Table~\ref{tab:examples_of_labeler}, and the numbers of negative and positive exams in the IU X-ray and MIMIC-CXR datasets are shown in Table~\ref{tab:output_dist_of_labeler}.

\subsection{Extending Model}\label{subsec:ext_model}

\begin{table*}[ht]
\centering
\begin{tabular}{c|c|cccccc}
\hline
\multirow{2}{*}{Dataset}  & \multirow{2}{*}{Model} & \multicolumn{6}{c}{NLG Metrics}       \\
                        &                        & BL-1 & BL-2 & BL-3 & BL-4 & CIDEr & RG-L \\ \hline
\multirow{4}{*}{IU X-Ray} & R2Gen~\cite{chen2020generating}  & 0.421     & 0.262     & 0.183     &  0.137    & 0.480   &  0.337    \\
                          & \cellcolor{Gray} w/ prior (ours)  &\cellcolor{Gray}\textbf{0.438}     &\cellcolor{Gray}\textbf{0.280}     & \cellcolor{Gray}\textbf{0.201}    & \cellcolor{Gray}\textbf{0.155}    & \cellcolor{Gray}\textbf{0.631}  &  \cellcolor{Gray}\textbf{0.351}  \\
                        & $M^2$Tr~\cite{cornia2020meshed} &0.400    &0.240     &0.159    &0.112     &0.300   &0.324     \\
                          & \cellcolor{Gray} w/ prior (ours) &\cellcolor{Gray}0.406      &\cellcolor{Gray}0.249      & \cellcolor{Gray}0.167     & \cellcolor{Gray}0.120     & \cellcolor{Gray}0.323   &  \cellcolor{Gray}0.330   \\ \hline 
\multirow{4}{*}{MIMIC-CXR} & R2Gen~\cite{chen2020generating}  &0.335      &0.206    &0.138      &0.100      &0.148    &0.278      \\
                          & \cellcolor{Gray} w/ prior (ours) &\cellcolor{Gray}0.342      &\cellcolor{Gray}0.222      & \cellcolor{Gray}\textbf{0.152}     & \cellcolor{Gray}\textbf{0.110}     & \cellcolor{Gray}\textbf{0.166}   &  \cellcolor{Gray}\textbf{0.301}   \\
                        & $M^2$Tr~\cite{cornia2020meshed}  &0.353      &0.211      &0.137      &0.094    &0.089    &0.262      \\
                          & \cellcolor{Gray} w/ prior (ours) &\cellcolor{Gray}\textbf{0.357}      &\cellcolor{Gray}\textbf{0.224}      & \cellcolor{Gray}0.151     & \cellcolor{Gray}0.108     & \cellcolor{Gray}0.101   &  \cellcolor{Gray}0.293   \\ \hline
\end{tabular}
\caption{Training results of the baseline models and models infused with prior information. The results of our approaches are shown in gray rows and the best metrics are bolded. All metrics are averaged over 3 runs. Full table with standard deviation is available in Table~\ref{tab:full_results}}
\label{tab:results}
\end{table*}

In this section, we describe how we integrate the comparison prior into existing models, such as R2Gen and $M^2$Tr, to generate more informative and comprehensive reports.

\paragraph{Generation Process}
The generation process of R2Gen and $M^2$Tr can be illustrated as shown in Figure~\ref{fig:concept}. It follows the following flow: input radiology images $X \rightarrow$ visual embedding $V \rightarrow$ latent representation $L \rightarrow$ output report $Y$. Initially, chest X-ray images $X$ are provided as inputs to the visual extractor, where $X$ consists of the frontal image $X_f$ and the lateral image $X_l$, represented as $X= \{X_f, X_l\}$. The visual extractor generates the visual embedding $V = \{ v_1, v_2, ..., v_S\}$, which comprises patch features $v_s \in \mathbb{R}^d$, with $d$ being the size of the feature vectors. Subsequently, $V$ undergoes multiple transformer layers in the encoder to obtain the latent representation $L = \{ l_1, l_2, ..., l_T\}$, where the latent feature vector is denoted as $l_t \in \mathbb{R}^f$. Finally, the decoder utilizes $L$ to generate the final output report $Y$.

\paragraph{Infusing Comparison Prior}
The comparison prior $P \in \mathbb{R}$ is generated from our rule-based labeler and it denotes a negative (0) or positive (1) label. We intend to incorporate the comparison prior into the existing data pipeline in such a way that the addition of the prior does not change the architecture or add any additional weights to train. Otherwise, it will become hard to measure the effect of comparison prior to the generative models. As a result, we added prior $P$ to both Visual Embedding $V$ and Latent Representation $L$ in the generation models shown in Figure~\ref{fig:concept}. The encoder should be given the prior information so that it can generate an appropriate intermediate representation. Furthermore, we also add $P$ on $L$ since the knowledge of $P$ could be weakened after deep transformer layers in the encoder. The decoder will generate the output report based on the latent representation conditioned on $P$. This whole process emulates the radiologists' examination with prior exams. Therefore, our new visual embedding $V_{new}$ and new latent representation $L_{new}$ can be calculated as follows:
\begin{equation}
    V_{new} = V \oplus P, \quad L_{new} = L \oplus P
\end{equation}
where $\oplus$ indicates element-wise summation. The strength of our method is that it is applicable to most existing transformer-based models and does not require an extra dataset or information.  

\section{Experiment}

\paragraph{Architecture}
To extract visual features, we utilize pretrained Convolutional Neural Networks (CNNs) such as DenseNet121 \cite{huang2017densely} and ResNet121 \cite{he2016deep}. Through empirical evaluation, we find that DenseNet performs better for our generation task, and thus, we select it as our base visual extractor. We adopt the structure of Meshed-Memory Transformer ($M^2$Tr) \cite{cornia2020meshed} and Relational Memory-driven Transformer (R2Gen) \cite{chen2020generating} to construct our encoder and decoder.

\paragraph{Datasets}
We evaluate our proposed methods on two widely-used English datasets for medical report generation tasks: IU X-ray \cite{demner2016preparing} and MIMIC-CXR \cite{johnson2019mimic}. The IU X-ray dataset is a publicly available radiology dataset that consists of 7,470 chest X-ray images and 3,955 radiology reports. Each report is paired with one frontal view image and, optionally, one lateral view image. MIMIC-CXR is a large chest radiograph database comprising 473,057 chest X-ray images and 206,563 reports. We train our model using intact data pairs, which include two images (frontal and lateral) and one report (Findings section). The datasets are divided into train, validation, and test sets following the data split described in \citet{chen2020generating}.

\paragraph{Training Details}
We first generate the comparison prior for each report using a rule-based labeler. Then, we train our model with the two images and the comparison prior as inputs, and the medical report as the output. We employ the Adam optimizer with an initial learning rate of 0.00005 for the visual extractor and 0.0001 for the encoder-decoder model. The learning rate gradually decreases at pre-defined steps. All experiments are conducted with 3 different seeds and a batch size of 16 on an "NVIDIA GeForce RTX 1080 Ti" GPU. Our code implementation is based on the publicly available codes from \citet{chen2020generating} and \citet{nooralahzadeh2021progressive}.

\paragraph{Evaluation Metrics}
We report general natural language generation (NLG) metrics, including BLEU \cite{papineni2002bleu}, CIDEr \cite{vedantam2015cider}, and ROUGE-L \cite{lin2004rouge}. These metrics are commonly used to evaluate the quality of generated text. BLEU measures the n-gram overlap between the generated text and the reference text, while CIDEr is based on cosine similarity between word embeddings and considers both unigrams and multi-word phrases. ROUGE-L evaluates the longest common subsequence between the generated text and the reference text. Including these metrics enables a quantitative comparison of the generated reports with the ground truth and previous models, providing insights into the performance of the proposed approach.

\paragraph{Results} 
In Table~\ref{tab:results}, we present the results of our proposed approach, which incorporates prior information into state-of-the-art NLG models, on two medical image report generation datasets: IU X-Ray and MIMIC-CXR. Our approach consistently outperforms the baselines across all NLG metrics, demonstrating its effectiveness in improving the quality of medical image reports generated by NLG models.

On the IU X-Ray dataset, our approach achieves an average improvement of 11.58\% and 4.49\% on all NLG metrics for R2Gen and $M^2$Tr models, respectively, compared to the baseline models. Notably, the CIDEr metric shows the highest improvement, with an increase of 31.46\% for R2Gen and 7.00\% for $M^2$Tr. This suggests that, as measured by CIDEr, our approach generates more diverse and contextually relevant captions, which align better with human judgments of quality than other metrics.

For the MIMIC-CXR dataset, our approach improves the R2Gen and $M^2$Tr models by 8.40\% and 9.62\% on all NLG metrics, respectively, compared to the previous models. The most significant improvement is observed in ROUGE-L, with an increase of 8.27\% for R2Gen and 11.83\% for $M^2$Tr. This indicates that our method produces more grammatically correct captions, which is particularly important in medical reports where language errors can have serious consequences.

We find that the highest order n-grams (i.e., n=3, 4) show the most significant improvements. This suggests that incorporating external prior information is especially beneficial for generating fluent and informative sentences that typically contain longer phrases and more complex structures.

Overall, our findings demonstrate that integrating external prior information can enhance the performance of existing NLG models for medical image reporting tasks, resulting in more informative and accurate medical reports. By incorporating additional domain-specific knowledge into the models, we are able to generate more precise and informative reports while minimizing computational overhead and training data requirements.

\section{Analysis}

In this section, we compare the ground truths, synthetic reports created by our proposed model, and two previously published models, R2Gen and $M^2$Tr, to assess the effectiveness of our model in generating concise and accurate reports without irrelevant or false priors.

Table~\ref{tab:compare_report_examples} in the Appendix provides examples of synthetic reports generated by each model and the corresponding ground truth for the same radiology image. The first two rows compare reports generated by R2Gen and our model with prior infusion. We observe that R2Gen generates false prior expressions, such as "compared to prior examination", "unchanged from prior", and "again unchanged", which refer to non-existent prior exams. In contrast, our model generates more concise and accurate reports without any prior expressions, resulting in higher performance in NLG metrics.

Similarly, the last two rows of Table~\ref{tab:compare_report_examples} in the Appendix compare reports generated by $M^2$Tr and our proposed model. $M^2$Tr produces reports with false prior expressions, such as "present on the previous exam" and "again noted", while our model avoids including any comparison phrases. Furthermore, reports that include prior expressions tend to be longer due to the additional explanations required for comparison. However, the report generation model does not actually have access to previous exams for comparison, rendering the inclusion of prior expressions irrelevant or misleading. As a result, our models can directly control these phrases by conditioning the generation through priors.

Overall, the synthetic reports generated by our proposed model are more concise and accurate compared to those generated by R2Gen and $M^2$Tr, as evidenced by the higher performance in NLG metrics. Our model achieves this by avoiding irrelevant or false prior expressions through a rule-based labeler and generating reports that contain only relevant and accurate information. These succinct and precise reports generated by our model will effectively assist radiologists in their practice.

\section{Limitations and Ethical Considerations}
Our proposed method has certain limitations and ethical considerations that merit discussion. The effectiveness of our approach heavily relies on the rule-based labeler. However, it is important to acknowledge that the labeler may not capture unseen patterns or variations, potentially limiting improvements in various evaluation metrics. Moreover, we were unable to conduct a comprehensive human evaluation of the rule-based labeler in this study due to resource constraints. Therefore, future work should include a detailed evaluation to assess its performance and address any potential limitations.

Collaboration with three radiologists at Kyoto University is a critical aspect of our work. The regular expressions designed in the rule-based labelers were validated through mutual confirmation by computer scientists and radiologists. However, it is essential to note that the radiologists involved in the collaboration primarily work in a Japanese hospital setting. This may introduce potential biases or patterns that are specific to the local context. Therefore, it is necessary to cross-check the performance of the rule-based labeler with radiologists from different regions and healthcare systems to ensure broader applicability and minimize any potential bias.

Regarding the datasets used in our study, we exclusively utilized publicly available datasets that are properly anonymized and de-identified, addressing privacy concerns. However, it is crucial to emphasize that if datasets containing comparison exams become available in the future, additional precautions must be taken to ensure that no personally identifiable information is inadvertently disclosed or used in a manner that could identify individual patients.

By acknowledging these limitations and ethical considerations, we aim to encourage future research and discussions in the field, driving advancements in radiology report generation while prioritizing patient privacy, accuracy, and fairness. These considerations will contribute to the development of robust and ethically sound approaches in radiology report generation.

\section{Conclusion}

In this study, we present a novel approach to generate medical reports from chest X-ray images, aiming to bridge the gap between radiologists' knowledge and the lack of prior information in generation models. To achieve this, we developed a rule-based labeler capable of extracting comparison priors from radiology reports in the IU X-ray and MIMIC-CXR datasets. These priors were subsequently integrated into state-of-the-art models for conditional report generation, allowing our approach to emulate the realistic diagnostic process of radiologists who possess prior information about patients.

Our experimental results demonstrate the superiority of our method over previous state-of-the-art models, as indicated by improved performance in terms of NLG metrics and a significant reduction in the occurrence of falsely referred prior exams. Through our analysis, we show that the incorporation of comparison priors leads to the generation of more accurate and concise reports, thereby holding great potential to enhance the quality and efficiency of medical report generation for chest X-ray images. Ultimately, this advancement benefits healthcare professionals and patients by providing more reliable and informative reports.

Furthermore, our work highlights the future potential of generating medical reports in an end-to-end fashion if a dataset containing all previous exams becomes available. The ability to leverage comprehensive prior information would further amplify the accuracy and effectiveness of medical report generation, paving the way for improved healthcare outcomes.


\bibliography{acl2023}
\bibliographystyle{acl_natbib}

\appendix
\begin{table*}
\centering
\begin{tabular}{p{0.3\linewidth}  p{0.3\linewidth}  p{0.3\linewidth}}
\hline
\multicolumn{1}{c}{Ground Truth} & \multicolumn{1}{c}{R2Gen \cite{chen2020generating}} & \multicolumn{1}{c}{R2Gen w/ prior (ours)}\\ \hline
{\small The cardiomediastinal silhouette is normal in size and contour. Low lung volumes without focal pneumothorax or large pleural effusion. Normal xxxx.}  & {\small The trachea is midline. The cardiomediastinal silhouette is normal and unchanged \textbf{compared to prior examination}. Densities overlying the heart xxxx are xxxx coronary artery stents. There are small round calcific densities in the bilateral lobes which are \textbf{unchanged from prior exam} and xxxx represent sequelae from old granulomatous disease.}   & {\small The cardiomediastinal silhouette is within normal limits for appearance. The thoracic aorta is tortuous. No focal areas of pulmonary consolidation. No pneumothorax. No large pleural effusion. Mild degenerative changes and osteopenia of the thoracic spine.}       \\ \hline

{\small Cardiomediastinal silhouette is within normal limits. No focal consolidation. There is right lower lobe scarring. No pneumothorax or large pleural effusion. Granulomas present. No acute bony abnormalities.}  & {\small The trachea is midline. The cardiomediastinal silhouette is normal. The superior thoracic spine is \textbf{again unchanged from prior}. Lucent pulmonary parenchyma is consistent appearance with emphysema and appears \textbf{unchanged from prior xxxx}. No evidence of pneumothorax. No focal airspace disease or pleural effusion. Vague density in the medial right lung apex most xxxx is.}   & {\small The cardiomediastinal silhouette is normal in size and contour. No focal pneumothorax or large pleural effusion. Negative for acute bone abnormality.}       \\ \hline

\multicolumn{1}{c}{Ground Truth} & \multicolumn{1}{c}{$M^2$Tr \cite{cornia2020meshed}} & \multicolumn{1}{c}{$M^2$Tr w/ prior (ours)}\\ \hline

{\small No change lung xxxx. xxxx opacities are present in the right lower lobe. No focal infiltrates. Heart and mediastinum are unremarkable. Aorta normal.}  & {\small Stable cardiomediastinal silhouette with normal heart mediastinal calcifications suggest a previous granulomatous process. Apical irregularities also \textbf{present on the previous exam} suggestive of scarring. No focal alveolar no definite pleural effusion seen. No typical findings of pulmonary edema. No pneumothorax.}   & {\small The heart is normal in size and contour. There is no mediastinal widening. The lungs are hyperexpanded. Scattered granuloma. No focal airspace disease. No large pleural effusion or pneumothorax. The xxxx are intact.}       \\ \hline

{\small The trachea is midline. Negative for pleural or focal airspace consolidation. The heart size is normal.}  & {\small The heart is top normal in size. The mediastinum is unremarkable. The lungs are hypoinflated but grossly clear. Significant degenerative changes of the xxxx are \textbf{again noted} bilaterally.}   & {\small The heart is normal in size. The mediastinum is unremarkable. The lungs are clear.}       \\ \hline

\end{tabular}
\caption{Ground truth report from IU X-ray (first column), reports generated by R2Gen and $M^2$Tr (second column), and reports generated by our model. Prior expressions are written in bold.}
\label{tab:compare_report_examples}
\end{table*}

\begin{table*}[ht]
\centering
\begin{tabular}{c|cccccc}
\hline
\multirow{2}{*}{Model} & \multicolumn{6}{c}{NLG Metrics (IU X-Ray)}       \\
                       & BL-1 & BL-2 & BL-3 & BL-4 & CIDEr & RG-L \\ \hline
R2Gen  & $0.421_{\pm0.001}$     & $0.262_{\pm0.003}$     & $0.183_{\pm0.004}$     &  $0.137_{\pm0.005}$    & $0.480_{\pm0.046}$   &  $0.337_{\pm0.005}$    \\
\cellcolor{Gray} w/ prior (ours)  &\cellcolor{Gray}$0.438_{\pm0.003}$     &\cellcolor{Gray}$0.280_{\pm0.002}$    & \cellcolor{Gray}$0.201_{\pm0.002}$   & \cellcolor{Gray}$0.155_{\pm0.002}$    & \cellcolor{Gray}$0.631_{\pm0.028}$  &  \cellcolor{Gray}$0.351_{\pm0.001}$  \\
$M^2$Tr  &$0.400_{\pm0.003}$    &$0.240_{\pm0.002}$     &$0.159_{\pm0.002}$    &$0.112_{\pm0.002}$     &$0.300_{\pm0.004}$   &$0.324_{\pm0.002}$     \\
\cellcolor{Gray} w/ prior (ours) &\cellcolor{Gray}$0.406_{\pm0.003}$      &\cellcolor{Gray}$0.249_{\pm0.002}$      & \cellcolor{Gray}$0.167_{\pm0.001}$     & \cellcolor{Gray}$0.120_{\pm0.001}$     & \cellcolor{Gray}$0.323_{\pm0.001}$   &  \cellcolor{Gray}$0.330_{\pm0.001}$   \\ \hline \hline 

\multirow{2}{*}{Model} & \multicolumn{6}{c}{NLG Metrics (MIMIC-CXR)}       \\
                       & BL-1 & BL-2 & BL-3 & BL-4 & CIDEr & RG-L \\ \hline
R2Gen  &$0.335_{\pm0.001}$      &$0.206_{\pm0.003}$    &$0.138_{\pm0.005}$      &$0.100_{\pm0.002}$      &$0.148_{\pm0.006}$    &$0.278_{\pm0.003}$      \\
\cellcolor{Gray} w/ prior (ours) &\cellcolor{Gray}$0.342_{\pm0.002}$      &\cellcolor{Gray}$0.222_{\pm0.002}$      & \cellcolor{Gray}$0.152_{\pm0.002}$     & \cellcolor{Gray}$0.110_{\pm0.001}$     & \cellcolor{Gray}$0.166_{\pm0.004}$   &  \cellcolor{Gray}$0.301_{\pm0.005}$   \\
$M^2$Tr   &$0.353_{\pm0.001}$      &$0.211_{\pm0.003}$      &$0.137_{\pm0.003}$      &$0.094_{\pm0.002}$    &$0.089_{\pm0.002}$    &$0.262_{\pm0.002}$      \\
\cellcolor{Gray} w/ prior (ours) &\cellcolor{Gray}$0.357_{\pm0.001}$      &\cellcolor{Gray}$0.224_{\pm0.001}$      & \cellcolor{Gray}$0.151_{\pm0.002}$     & \cellcolor{Gray}$0.108_{\pm0.003}$     & \cellcolor{Gray}$0.101_{\pm0.001}$   &  \cellcolor{Gray}$0.293_{\pm0.003}$   \\ \hline
\end{tabular}
\caption{Training results of the baseline models and models infused with prior information. Upper table is the results from IU X-Ray and Lower table is the results from MIMIC-CXR. The results of our approaches are shown in gray rows. All metrics are averaged over 3 runs (mean $\pm$ standard deviation).}
\label{tab:full_results}
\end{table*}

%\section{Example Appendix}
%\label{sec:appendix}

%This is a section in the appendix.

\end{document}